\pdfoutput=1

\documentclass[11pt]{article}

\usepackage{EMNLP2022}

\usepackage{times}
\usepackage{latexsym}

\usepackage[T1]{fontenc}

\usepackage[utf8]{inputenc}

\usepackage{microtype}

\usepackage{inconsolata}

\usepackage[utf8]{inputenc} 
\usepackage[T1]{fontenc}    
\usepackage{hyperref}       
\usepackage{url}            
\usepackage{booktabs}       
\usepackage{amsfonts}       
\usepackage{amsmath}
\usepackage{nicefrac}       
\usepackage{microtype}      
\usepackage{xcolor}         
\usepackage{amssymb}
\usepackage{graphics}
\usepackage{listings}
\usepackage{xcolor}
\usepackage{csquotes}
\usepackage{cleveref}
\usepackage{pgfplots}
\usepackage{pgfplotstable}
\usepackage{tikz}
\usepackage{float}
\usepackage{pifont}
\newtheorem{definition}{Definition}

\newcommand{\cmark}{\ding{51}}%
\newcommand{\xmark}{\ding{55}}%

\DeclareMathOperator*{\argmin}{arg\,min}

\newcommand{\sbpm}[1]{\scalebox{0.7}{$\pm$#1}}

\title{Less is More: Parameter-Free Text Classification with Gzip}

\author{Zhiying Jiang, Matthew Y.R. Yang, Mikhail Tsirlin, Raphael Tang, \and Jimmy Lin \\[1ex]
  David R. Cheriton School of Computer Science \\
  University of Waterloo \\[1ex]
  \texttt{\{zhiying.jiang, m259yang, mtsirlin, r33tang, jimmylin\}@uwaterloo.ca}}

\begin{document}

\maketitle

\begin{abstract}

Deep neural networks (DNNs) are often used for text classification tasks as they usually achieve high levels of accuracy. However, DNNs can be computationally intensive with billions of parameters and large amounts of labeled data, which can make them expensive to use, to optimize and to transfer to out-of-distribution (OOD) cases in practice. In this paper, we propose a non-parametric alternative to DNNs that's easy, light-weight and universal in text classification: a combination of a simple compressor like gzip with a $k$-nearest-neighbor classifier. Without any training, pre-training or fine-tuning, our method achieves results that are competitive with non-pretrained deep learning methods on six in-distributed datasets. It even outperforms BERT on all five OOD datasets, including four low-resource languages. Our method also performs particularly well in few-shot settings where labeled data are too scarce for DNNs to achieve a satisfying accuracy.
\end{abstract}

\section{Introduction}

Text classification, as one of the most fundamental tasks in natural language processing (NLP), has improved substantially with the help of neural networks~\cite{li2022survey}.
However, most neural networks are data hungry, the degree of which increases with the number of parameters.
They also have many hyperparameters that must be carefully tuned for different datasets, and the preprocessing of text data (e.g., tokenization, stop word removal) must be tailored to the specific model and dataset. Despite their ability to capture latent correlations and recognize implicit patterns~\cite{lecun2015deep} complex deep neural networks may be overkill for simple tasks such as text classification.
For example, \citet{adhikari2019rethinking} find that a simple long short-term memory network (LSTM; \citealp{hochreiter1997long}) with appropriate regularization can achieve competitive results. \citet{shen2018baseline} further show that even word-embedding-based methods can achieve results comparable to convolutional neural networks (CNNs) and recurrent neural networks (RNNs). 

In this paper, we propose a simple, lightweight and universal alternative to DNNs for text classification that combines a lossless compressor with a $k$-nearest-neighbor classifier. It’s simple because it doesn’t require any pre-processing or training. It’s lightweight in that it achieves results competitive to DNN methods without the need of parameters or GPU resource. It’s universal as compressors are data-type agnostic, non-parametric methods do not bring inductive bias by the training procedure and it performs well on out-of-distribution (OOD) cases, where datasets are unseen by the model during pre-training or training stage. 

Lossless compressors aim to represent information using as few bits as possible by assigning shorter codes to symbols with higher probability. 
The intuition of using compressor for classification is that (1) compressors are good at capturing regularity; (2) objects from the same category share more regularity than those that aren't. For example, $x_1$ below belongs to the same category as $x_2$ but belongs to a different category from $x_3$. If we use $C(\cdot)$ to represent compressed length, we will find $C(x_1x_2)-C(x_1)<C(x_1x_3)-C(x_1)$ where $C(x_1x_2)$ means the compressed length of concatenation of $x_1$ and $x_2$. In other words, $C(x_1x_2)-C(x_1)$ can be interpreted as how many bytes we can save to encode $x_2$ if we know $x_1$. 
\begin{quote}
    \textit{$x_1$ = Japan's Seiko Epson Corp. said Wednesday it has developed a 12-gram flying microrobot, the world's lightest.}
    
    \textit{$x_2$ = The latest tiny flying robot that could help in search and rescue or surveillance has been unveiled in Japan.}
    
    \textit{$x_3$ = Michael Phelps won the gold medal in the 400 individual medley and set a world record in a time of 4 minutes 8.26 seconds.}
    
\end{quote}
This simple intuition can be formalized as a distance metric derived from Kolmogorov complexity~\cite{kolmogorov1963tables} which will be discussed in detail in~\Cref{sec:ncd}.

Our contributions are as follows: 
(1) We propose a parameter-free method that achieves results comparable to non-pretrained neural network models that have millions of parameters on six out of seven in-distributed datasets;
(2) We demonstrate that our method outperforms non-pretrained neural networks in few-shot settings when labeled data is extremely limited;
(3) We show that our method outperforms pre-trained models on out-of-distributed datasets, under both full and few-shot settings;
(4) We find that, as a universal baseline, our method is particularly effective for datasets that are easily compressible.

\section{Related Work}

\subsection{Compressor-Based Text Classification}
\label{sec:ctc}
Compressor-based distance metrics have been used mainly for plagiarism detection~\cite{chen2004shared}, clustering~\cite{vitanyi2009normalized} and classifying time series data~\cite{chen1999compression, keogh2004towards}.

Several previous works explore methods using a compressor-based distance metric for text classification: \citet{li2004similarity} applies it to language identification as language are different in length by nature (e.g., \textit{addresses} <English>, \textit{adressebok} <Norwegian>, \textit{adressekartotek} <Danish>); \citet{khmelev2003repetition} uses it for authorship categorization; \citet{frank2000text,teahan2003using} utilize Prediction by Partial Matching (PPM) for topic classification. PPM, a text compression scheme utilizing language modeling, estimates the cross entropy between the probability distribution built on class $c$ and the document $d$: $H_c(d)$. The intuition is that the lower the cross entropy is, the more likely that $d$ belongs to $c$. 

Summarized in~\citet{russell2010artificial}, the procedure of using compressor to estimate $H_c(d)$ is that: (1) for each class $c$, concatenate all samples $d_c$ in the training set belonging to $c$; (2) compress $d_c$ as one long document to get the compressed length $C(d_c)$; (3) concatenate the given test sample $d_u$ with $d_c$ and compress to get $C(d_cd_u)$; (4) the predicted class is $\argmin_c C(d_cd_u)-C(d_c)$. 

The major drawback of this method is that concatenating all training documents in one class makes it hard to take full advantage of large training set, as compressors like \textit{gzip} has a limited size of sliding window, which is responsible for ``how much'' the compressor can look back to find repeated patterns.  \citet{marton2005compression} further investigate the distance metric $C(d_c^{(i)}d_c)-C(d_c^{(i)})$ where $d_c^{(i)}$ is a \textit{single} document belonging to class $c$.
\citet{coutinho2015text, kasturi2022text} focus on improving representations based on compressor to improve the classification accuracy.

To the best of our knowledge, all the previous work use relatively small datasets like 20News and Reuters-10. 
There is neither a comparison between compressor-based methods and deep learning methods nor any comprehensive study on large-sized datasets.

\subsection{DNN-Based Text Classification}
The deep learning methods used for text classification can be divided into two: transductive learning, represented by Graph Convolutional Networks (GCN)~\cite{yao2019graph}, and inductive learning, where both recurrent neural networks (RNN) and convolutional neural networks (CNN) are main forces. We focus on inductive learning in this paper as transductive learning assumes the test dataset is presented during the training. 

\citet{zhang2015character} first use the character-based CNN with millions of parameters for text classification. \citet{conneau2017very} extend the idea with more layers. 
Along the line of RNNs,~\citet{kawakami2008supervised} introduce a method that uses LSTMs~\cite{hochreiter1997long} to learn the sequential information for classification. To better capture the important information regardless of its position in the sentence,~\citet{wang2016attention} incorporate the attention mechanism into the relation classification.~\citet{yang2016hierarchical} include a hierarchical structure for sentence-level attention. 

As the number of parameters and the complexity of models increase, \citet{joulin2017bag} start to explore the possibility of using simple linear model with a hidden layer coping with $n$-gram features and hierarchical softmax to improve efficiency. 

The status quo of classification is further changed by the prevalence of pre-trained models like BERT~\cite{kenton2019bert}, with thousands of millions of parameters pre-trained on corpus containing billions of words. BERT can achieve the state of the art on numerous tasks including text classification~\cite{adhikari2019docbert} with just some fine-tunings. Built on BERT,~\citet{reimers2019sentence} calculate semantic similarity between pairs of sentences efficiently by using a siamese network architecture and fine-tuning on multiple NLI datasets~\cite{bowman2015large, williams-etal-2018-broad}. 

\section{Our Approach}
\label{sec:ncd}
Kolmogorov complexity $K(x)$ characterizes the length of the shortest binary program that can generate $x$. $K(x)$ is theoretically the ultimate lower bound for information measurement. Given this notion of information measurement, how, do we compare information content between two objects? 
To this end, \citet{bennett1998information} define \textit{information distance} $E(x,y)$ as the length of the shortest binary program that converts $x$ to $y$:
\begin{align}
            E(x,y) & = \max\{K(x|y), K(y|x)\} \\
    & = K(xy) - \min \{K(x), K(y)\}
\end{align}

However, $E(x,y)$ is not computable as Kolmogorov complexity is incomputable, and absolute distance makes comparison among objects hard.~\citet{li2004similarity} proposes a normalized and computable version of information distance, \textit{Normalized Compression Distance} (NCD), utilizing compressed length $C(x)$ to approximate Kolmogorov complexity $K(x)$. Formally, it's defined as follows
(detailed derivation is shown in~\Cref{appx:ncd}):
\begin{equation}
    NCD(x,y) = \frac{C(xy)-\min \{C(x), C(y)\}}{\max\{C(x), C(y)\}}
\end{equation}
The intuition behind using compressed length is that the length of $x$ that has been maximally compressed by a compressor is close to $K(x)$. The higher the compression ratio, the closer $C(x)$ is to $K(x)$.
Our main experiment results use \textit{gzip} as the compressor, thus, $C(x)$ means the length of $x$ after compressed by \textit{gzip}. $C(xy)$ is the compressed length of concatenation of $x$ and $y$. With the distance matrix NCD provides, we can then use $k$-nearest-neighbor to classify.

Our method can be implemented with fifteen lines of Python code below, whose input is \textit{training\_set}, \textit{test\_set}, both of which consist of an array of \textit{(text, label)}, and \textit{k}:

\definecolor{codegreen}{rgb}{0,0.6,0}
\definecolor{codegray}{rgb}{0.5,0.5,0.5}
\definecolor{codepurple}{rgb}{0.58,0,0.82}
\definecolor{backcolour}{rgb}{0.95,0.95,0.92}

\lstdefinestyle{mystyle}{
  backgroundcolor=\color{white}, commentstyle=\color{codegreen},
  keywordstyle=\color{magenta},
  numberstyle=\tiny\color{codegray},
  stringstyle=\color{codepurple},
  basicstyle=\ttfamily\footnotesize,
  breakatwhitespace=false,         
  breaklines=true,                 
  captionpos=b,                    
  keepspaces=true,                 
  numbers=left,                    
  numbersep=5pt,                  
  showspaces=false,                
  showstringspaces=false,
  showtabs=false,                  
  tabsize=2
}

\lstset{style=mystyle}

\begin{lstlisting}[language=Python, caption=Python Code for Text Classification with Gzip]
import gzip
import numpy as np

for (x1, _) in test_set:
    Cx1 = len(gzip.compress(x1.encode()))
    distance_from_x1 = []
    for (x2, _) in training_set:
        Cx2 = len(gzip.compress(x2.encode())
        x1x2 = " ".join([x1, x2])
        Cx1x2 = len(gzip.compress(x1x2.encode())
        ncd = (Cx1x2 - min(Cx1,Cx2)) / max(Cx1, Cx2)
        distance_from_x1.append(ncd)
    sorted_idx = np.argsort(np.array(distance_from_x1))
    top_k_class = training_set[sorted_idx[:k], 1]
    predict_class = max(set(top_k_class), key=top_k_class.count)

\end{lstlisting}

\section{Experiments}

\subsection{Datasets}
We choose this diverse basket of datasets to investigate the effects of the number of training samples, the number of classes, the length of the text and the difference in distribution on accuracy. The details of each dataset's statistics are listed in~\Cref{tab:dataset}. Previous works on text classification have two disjoint preferences when choosing evaluation datasets: CNN and RNN-based methods favor large scale datasets (AG News, SogouNews, DBpedia, YahooAnswers) for evaluation, whereas transductive methods like graph convolutional neural network focus on datasets with smaller training sets (20News, Ohsumed, R8, R52)~\cite{li2022survey}. We include datasets on both sides in order to investigate how our method performs with both abundant training samples and limited ones. Apart from the variation of the dataset sizes, we also take the effects of number of classes into consideration by intentionally including datasets like R52 to evaluate our the performance on datasets with large number of classes. Previous work~\cite{marton2005compression} show that the length of text also affects the accuracy of compressor-based methods so we present the statistics in~\Cref{tab:dataset} as well.
Except for SogouNews, we also include other four out-of-distributed datasets --- Kinyarwanda news, Kirundi news, Filipino dengue and Swahili news to further evaluate our method's robustness.

\subsection{Baselines}

\begin{table*}[h]
    \centering
    \begin{small}
    
    \begin{tabular}{c|c|c|c|c|c|c}
       Dataset & \#Training & \#Test & \#Classes & Avg\#Words & Avg\#Chars & \#Vocab \\
       \toprule
        AG News & 120,000 & 7,600 & 4 & 43.9 & 236.4 & 128,349\\
        DBpedia & 560,000 & 70,000 & 14 & 53.7 & 301.3 & 1,031,601\\
        YahooAnswers & 1,400,000 & 60,000 & 10 & 107.2 & 520.8 & 1,554,607\\
        20News & 11,314 & 7,532 & 20 & 406.02 & 1902.5 & 277,330\\
        ohsumed & 3,357 & 4,043 & 23 & 212.1 & 1273.2 & 55,142\\
        R8 & 5,485 & 2,189 & 8 & 102.4 & 586.8 & 23,584\\
        R52 & 6,532 & 2,568 & 52 & 109.6 & 631.4 & 26,283\\
        KinyarwandaNews & 17,014 & 4,254 & 14 & 232.3 & 1872.3 & 240,366 \\
        KirundiNews & 3,689 & 923 & 14 & 210.2 & 1721.5 & 63,143\\
        DengueFilipino & 4,015 & 500 & 5 & 10.1 & 62.7 & 12,819\\
        SwahiliNews & 22,207 & 7,338 & 6 & 327.0 & 2196.5 & 569,603 \\
        SogouNews & 450,000 & 60,000 & 5 & 589.4 & 2780.0 &   610,908\\
        \bottomrule
    \end{tabular}
    \caption{Details of datasets used for evaluation.}
    \label{tab:dataset}
    \end{small}
\end{table*}

\begin{table*}[h]
    \centering
    \begin{small}
    
    \begin{tabular}{c|c|c|c|c|c}
       Model & \#Param & Pre-training & Training & External Data & Pre-Process \\
       \toprule
        TFIDF+LR & 260,000 & \xmark & \cmark & \xmark & tok+tfidf+dict (+lower) \\
        LSTM & 5,190,000 & \xmark & \cmark & \xmark & tok+dict (+wv+lower+pad) \\
        Bi-LSTM+Attn & 8,210,000 & \xmark & \cmark & \xmark & tok+dict (+wv+lower+pad) \\
        HAN & 29,700,000 & \xmark & \cmark & \xmark & tok+dict (+wv+lower+pad) \\
        charCNN & 2,700,000 & \xmark & \cmark & \xmark & dict (+lower+pad) \\
        textCNN & 30,700,000 & \xmark & \cmark & \xmark & tok+dict (+wv+lower+pad) \\
        RCNN & 18,800,000 & \xmark & \cmark & \xmark & tok+dict (+wv+lower+pad) \\
        VDCNN & 13,700,000 & \xmark & \cmark & \xmark & dict (+lower+pad) \\
        fasttext & 8,190,000 & \xmark & \cmark  & \xmark & tok+dict (+lower+pad+ngram) \\
        BERT & 109,000,000 & \cmark & \cmark & \cmark & tok+dict+pe (+lower+pad) \\
        W2V & 0 & \cmark & \xmark & \xmark & tok+dict (+lower) \\
        SentBERT & 0 & \cmark & \xmark & \cmark & tok+dict (+lower) \\
        TextLength & 0 & \xmark & \xmark & \xmark & \xmark \\
        gzip & 0 & \xmark & \xmark & \xmark & \xmark \\

        \bottomrule
    \end{tabular}
    \caption{Models used for comparison and their number of parameters; whether they are pre-trained; whether data augmentation is used and whether pre-processing is needed.}
    \label{tab:param}
    \end{small}
\end{table*}

We compare our result with (1) neural network methods that require training and (2) zero-training methods that use the $k$NN classifier directly, with or without pre-training. 
Specifically, we choose mainstream architectures for text classification, like logistic regression, fasttext~\cite{joulin2017bag}, RNNs with or without attention (vanilla LSTM~\cite{hochreiter1997long}, bidirectional LSTMs~\cite{schuster1997bidirectional} with attention~\cite{wang2016attention}, hierarchical attention networks~\cite{yang2016hierarchical}), CNNs (character CNNs~\cite{zhang2015character}, recurrent CNNs~\cite{lai2015recurrent}, very deep CNNs~\cite{conneau2017very}) and BERT~\cite{devlin2019bert}~\cite{adhikari2019docbert}. We also include three other zero-training methods: word2vec (W2V)~\cite{mikolov2013efficient}, pre-trained sentence BERT~\cite{reimers2019sentence}, and the length of the instance, all using a $k$NN classifier. To prevent the class from being predicted based on text length, we evaluate a baseline where the instance text length is used as the only input into a $k$NN classifier.
We call this baseline the TextLength method.

We present model statistics and trade-offs in~\Cref{tab:param}.
Since the number of classes, the vocabulary size, and the dimensions affect the number of parameters, we estimate the model size using AGNews.
This dataset has a relatively small vocabulary size and number of classes, hence making the estimation of the lower bound out of the studied datasets. Some methods require pre-training either on the target dataset or on other external datasets. Most neural networks require pre-processing like tokenization (``tok''), building vocabulary dictionaries and mapping tokens (``dict''), using pre-trained word2vec (``wv''), lowercasing the words (``lower'') and padding the sequence to a certain length (``pad''). Other model-specific pre-processing includes adding extra bag of n-grams (``ngram'') for \textit{fasttext} and using positional embedding (``pe'') for \textit{BERT}.

\begin{table*}[h]
    \centering
    
    \small
    \begin{tabular}{c|c|c|c|c|c|c|c}
        Model/Dataset & AGNews & DBpedia & YahooAnswers & 20News & Ohsumed & R8 & R52 \\
        \toprule
        \multicolumn{8}{c}{Training Required} \\
        \hline
        TFIDF+LR & 0.898 & 0.982 & 0.715 & 0.827 & 0.549 & 0.949 & 0.874  \\
        \hline
        LSTM & 0.861 & 0.985 & 0.708 & \underline{0.657} & 0.411 & 0.937 & 0.855   \\
        \hline
        Bi-LSTM+Attn & \underline{0.917} & 0.986 & 0.732 & 0.588 & 0.271 & 0.868 & 0.693  \\
        \hline
        HAN & 0.896 & 0.986 & 0.745 & 0.646 & 0.462 & 0.960 & 0.914   \\
        \hline
        charCNN & 0.914 & 0.986 & 0.712 & 0.401 & 0.269 & 0.823 & 0.724   \\
        \hline
        textCNN & 0.817 & 0.981 & 0.728 & 0.751 & 0.570 & \underline{0.951} & \underline{0.895}   \\
        \hline
        RCNN & 0.912 & 0.984 & 0.702 & 0.716 & \underline{0.472} & 0.810 & 0.773   \\
        \hline
         VDCNN & 0.913 & 0.987 & 0.734 & 0.491 & 0.237 & 0.858 & 0.750  \\
         \hline
         fasttext & 0.911 & 0.978 & 0.702 & 0.690 & 0.218 & 0.827 & 0.571 \\
         
         \hline
         BERT & \textbf{0.944} & \textbf{0.992} & \textbf{0.768} & \textbf{0.868} & \textbf{0.741} & \textbf{0.982} & \textbf{0.960}   \\

         \toprule
         \multicolumn{8}{c}{Zero Training} \\
        \hline
         W2V & 0.892 & \underline{\textbf{0.961}} & 0.689 & 0.460 & 0.284 & 0.930 & 0.856  \\
         \hline
         SentBERT & \textbf{0.940} & 0.937 & \textbf{0.782} & \textbf{0.778} & \textbf{0.719} & \textbf{0.947} & \textbf{0.910}  \\
        \toprule
         \multicolumn{8}{c}{Zero Training \& Zero Pre-Training} \\
        \hline
        TextLength & 0.275 & 0.093 & \underline{0.105} & 0.053 & 0.090 & 0.455 & 0.362 \\
        \hline
         gzip (ours)  & \textbf{0.937} & \textbf{0.970} & \textbf{0.638} & \textbf{0.685} & \textbf{0.521} & \textbf{0.954} & \textbf{0.896}  \\
         \hline
    \end{tabular}
    \caption{Test accuracy with each section's best results bolded, and best results beaten by \textit{gzip} underlined.}
    \label{tab:full}
\end{table*}

\subsection{Result on In-Distributed Datasets}
We train all baselines on eight datasets (training details are in~\Cref{appx:train}). The result of using the full training sets are shown in~\Cref{tab:full}. As we can see, our method performs surprisingly well on AG News, R8 and R52. For AG News, fine-tuning BERT achieves the best performance among all methods, and \textit{gzip}, with no pretraining, achieves competitive result, within 0.007 points of BERT. 
The accuracy of \textit{gzip} on DBpedia is about $1\%$ lower than other neural network methods. For YahooAnswers, the accuracy of \textit{gzip} is about $7\%$ lower than the average neural methods. This may due to the fact that the vocabulary size of YahooAnswers is large, making it hard for the compressor to compress (detailed discussion is in~\Cref{sec:ana}). 

Starting from 20News dataset, the training size becomes smaller, where non-pretrained deep learning models are thought to be less advantageous. On the 20News dataset, pre-trained methods achieve the best result and \textit{gzip}'s accuracy is in the middle. Ohsumed is a dataset containing paper abstracts in the medical domain, aimed at categorizing 23 cardiovascular diseases. On Ohsumed, \textit{gzip} is lower than textCNN, BERT, SentBERT, competitive to LR and higher than others. 
For R8, \textit{gzip} has the third highest accuracy, only lower than HAN and BERT. For R52, \textit{gzip} ranks the fourth, surpassed by HAN, BERT and SentBERT. 

Overall, BERT-based models are robust even when the size of training samples are small, but do not excel when the dataset is out of distributed of the pre-training corpus (e.g., SogouNews). Character-based models like charCNN and VDCNN perform badly when the training data is small and the vocabulary size is large (e.g., 20News). The advantage of word-based models is non-obvious when the training data is small either, but they are better at handling big vocabulary size. They are also inferior to character-based models when classifying corpus that are not English, similar to BERT-based models. Logistic regression with TFIDF features, although doesn't achieve the best on any dataset, is very robust to the size of the dataset. The result of TextLength is close to random guess on all but R8 and R52, showing that the distribution of length doesn't reflect the information of class in other six datasets, indicating the compressed length information used in NCD does not benefit from the length distribution of different classes.

\begin{table}[h]
    \centering
    \begin{tabular}{c|c|c}
        Dataset & average & gzip \\
        \hline
        AGNews & 0.901 & \textbf{0.937} \\
        DBpedia & 0.978 & 0.970 \\
        YahooAnswers & 0.726 & 0.638 \\
        20News & 0.656 & \textbf{0.685} \\
        Ohsumed & 0.433 & \textbf{0.521} \\
        R8 & 0.903 & \textbf{0.954} \\
        R52 & 0.815 & \textbf{0.896} \\
        \hline
    \end{tabular}
    \caption{Test accuracy comparison between the average of all baseline models (excluding TextLength) and \textit{gzip}. }
    \label{tab:cmp_avg}
\end{table}

\textit{gzip} does not perform well on extremely large dataset (e.g., YahooAnswers), but are competitive on medium and small-size datasets. Performance-wise, the only non-preptrained deep learning model that's competitive to \textit{gzip} is HAN, who surpass \textit{gzip} on 50\% datasets and still achieve relatively high accuracy when it's beaten by \textit{gzip}, unlike textCNN. The difference is that \textit{gzip} doesn't require training.

We list the average of all baseline models' test accuracy (except TextLength for its extremely low accuracy) in~\Cref{tab:cmp_avg}. We can see our method is either higher or close to the average on all but YahooAnswers.

\subsection{Result on Out-Of-Distributed Datasets}
\begin{table*}[]
    \centering
    \resizebox{\textwidth}{!}{%
    \begin{tabular}{c|c|c|c|c|c|c|c|c|c|c}
       Model/Dataset & \multicolumn{2}{c|}{KinyarwandaNews}  & \multicolumn{2}{c|}{KirundiNews} & \multicolumn{2}{c|}{DengueFilipino} & \multicolumn{2}{c|}{SwahiliNews} & \multicolumn{2}{c}{SogouNews} \\
        \hline
        Shot\# & Full & 5-shot & Full & 5-shot & Full & 5-shot & Full & 5-shot & Full & 5-shot \\
        \hline
        BERT & 0.838 & 0.240$\pm$0.060 & 0.879 & 0.386$\pm$0.099 & 0.979 & 0.409$\pm$0.058 & 0.897 & 0.396$\pm$0.096 & 0.952 & 0.221$\pm$0.041\\
        \hline
        mBERT & 0.835 & 0.229$\pm$0.066 & 0.874 & 0.324$\pm$0.071 & 0.983 & 0.465$\pm$0.048 & 0.906 & 0.558$\pm$0.169 & 0.953 & 0.282$\pm$0.060 \\
        \hline
        gzip (ours) & \textbf{0.891} & \textbf{0.458$\pm$0.065} & \textbf{0.905} & \textbf{0.541$\pm$0.056} & \textbf{0.998} & \textbf{0.652$\pm$0.048} & \textbf{0.927} & \textbf{0.627$\pm$0.072} & \textbf{0.975} & \textbf{0.649$\pm$0.061}\\
        \hline
    \end{tabular}
    }
    \caption{Test accuracy on out-of-distributed datasets with 95\% confidence interval over five trials in five-shot setting.}
    \label{tab:ood}
\end{table*}

Generalizing to Out-Of-Distributed datasets have always been a challenge in machine learning. Even with the success of pre-trained models, this problem is not alleviated. In fact, \citet{yu2021empirical} have shown that improved in-distributed accuracy on pre-trained models may lead to poor OOD performance in image classification.
In order to compare our method with pre-trained models on text classification, we choose five datasets that are unseen in BERT's pre-trained corpus. Specifically, we use Kinyarwanda news, Kirundi news, Filipino dengue, Swahili news and Sogou news. Those datasets are chosen to have Roman script which means they have a very similar alphabet as English. For example, Swahili has the same vowels as English but doesn't have \texttt{q,x} as consonants; Sogou news only have Pinyin -- a phonetic romanization of Chinese. Therefore, those datasets can be viewed as permutation of English alphabets. 

We use BERT pre-trained on English and BERT pre-trained on 104 languages (mBERT). We can see that on languages that mBERT has been pre-trained on (Kinyarwanda, Kirundi or Pinyin), mBERT has lower accuracy than BERT in both full-data setting and few-shot setting. On Filipino and Swahili, mBERT has much higher accuracy than BERT especially in few-shot setting. However, on all five datasets, our method outperform both BERT and mBERT by large margin without any pre-training or fine-tuning. 

This shows the robustness of our method facing the OOD datasets. Our method is universal in a way that it is designed to handle unseen datasets as compressor is data-type-agnostic and non-parametric methods do not bring inductive bias induced by the training procedure.

\begin{figure*}[th]
\centering
\begin{tikzpicture}[scale = 0.28]
\begin{axis}[
width=1\textwidth,
height=0.80\textwidth,
legend cell align=left,
font=\LARGE,
axis y line*=left,
xmin=5, xmax=100,domain=1:10,
ymin=0.0, ymax=1.0,
every axis plot/.append style={thick},
xtick={5,10,50,100},
ytick={0.0, 0.2, 0.4, 0.6, 0.8, 1.0},
legend pos=south east,
xmajorgrids=true,
ymajorgrids=true,
tick label style={font=\LARGE},
legend style={nodes={scale=1.2, transform shape}},
xlabel style={font = \huge, yshift=0ex},
xlabel=\# of shots,
ylabel= Test Accuracy,
ylabel style={font = \huge, yshift=0ex}]

\addplot[
  black, mark=triangle*, red, mark options={scale=3},
  error bars/.cd, 
    error bar style={line width=2pt,solid},
  error mark options={line width=2pt,mark size=4pt,rotate=90},
    y fixed,
    y dir=both, 
    y explicit
] table [x=x, y=y,y error=error, col sep=comma] {
    x,    y,       error
    5, 0.273, 0.021
    10, 0.329, 0.036
    50, 0.550, 0.008
    100, 0.684, 0.010

};
\addlegendentry{fasttext}

\addplot[
  black, mark=triangle*, blue, mark options={scale=3},
  error bars/.cd,
    error bar style={line width=2pt,solid},
  error mark options={line width=2pt,mark size=4pt,rotate=90},
  y fixed,
  y dir=both,
  y explicit
] table [x=x, y=y, y error=error, col sep=comma]{
  x, y, error
  5, 0.269, 0.022
  10, 0.331, 0.028
  50, 0.549, 0.028
  100, 0.665, 0.019
};
\addlegendentry{Bi-LSTM+Attn}

\addplot[
  black, mark=triangle*, cyan, mark options={scale=3},
  error bars/.cd,
    error bar style={line width=2pt,solid},
  error mark options={line width=2pt,mark size=4pt,rotate=90},
  y fixed,
  y dir=both,
  y explicit
] table [x=x, y=y, y error=error, col sep=comma]{
  x, y, error
  5, 0.274, 0.024
  10, 0.289, 0.020
  50, 0.340, 0.073
  100, 0.548, 0.031 
};
\addlegendentry{HAN}

\addplot[
  black, mark=triangle*, orange, mark options={scale=3},
  error bars/.cd,
    error bar style={line width=2pt,solid},
  error mark options={line width=2pt,mark size=4pt,rotate=90},
  y fixed,
  y dir=both,
  y explicit
] table [x=x, y=y, y error=error, col sep=comma]{
  x, y, error
  5, 0.388, 0.186
  10, 0.546, 0.162
  50, 0.531, 0.272
  100, 0.395, 0.089
};
\addlegendentry{W2V}

\addplot[
  black, mark=triangle*, green, mark options={scale=3},
  error bars/.cd,
    error bar style={line width=2pt,solid},
  error mark options={line width=2pt,mark size=4pt,rotate=90},
  y fixed,
  y dir=both,
  y explicit
] table [x=x, y=y, y error=error, col sep=comma]{
  x, y, error
  5, 0.716, 0.032
  10, 0.746, 0.018
  50, 0.818, 0.008
  100, 0.829, 0.004
};
\addlegendentry{SentBERT}

\addplot[
  black, mark=triangle*, olive, mark options={scale=3},
    error bars/.cd,
    error bar style={line width=2pt,solid},
  error mark options={line width=2pt,mark size=4pt,rotate=90},
  y fixed,
  y dir=both,
  y explicit
] table [x=x, y=y, y error=error, col sep=comma]{
  x, y, error
  5, 0.803, 0.026
  10, 0.819, 0.019
  50, 0.869, 0.005
  100, 0.875, 0.005
};
\addlegendentry{BERT}

\addplot[
  black, mark=*, black, mark options={scale=3},
  error bars/.cd,
    error bar style={line width=2pt,solid},
  error mark options={line width=2pt,mark size=4pt,rotate=90},
  y fixed,
  y dir=both,
  y explicit
] table [x=x, y=y, y error=error, col sep=comma]{
  x, y, error
  5, 0.587, 0.048
  10, 0.610, 0.034
  50, 0.699, 0.017
  100, 0.741, 0.007
};
\addlegendentry{gzip}

\end{axis}
\end{tikzpicture}%
\qquad
\begin{tikzpicture}[scale=0.28]
\begin{axis}[
width=1\textwidth,
height=0.80\textwidth,
legend cell align=left,
font=\LARGE,
axis y line*=left,
xmin=5, xmax=100,domain=1:10,
ymin=0.0, ymax=1.0,
every axis plot/.append style={thick},
xtick={5,10,50,100},
ytick={0.0, 0.2, 0.4, 0.6, 0.8, 1.0},
legend pos=south east,
xmajorgrids=true,
ymajorgrids=true,
legend style={nodes={scale=1.2, transform shape}},
tick label style={font=\LARGE},
xlabel style={font = \huge, yshift=0ex},
xlabel=\# of shots,
ylabel= Test Accuracy,
ylabel style={font = \huge, yshift=0ex}]

\addplot[
  black, mark=triangle*, red, mark options={scale=3},
  error bars/.cd, 
      error bar style={line width=2pt,solid},
  error mark options={line width=2pt,mark size=4pt,rotate=90},
    y fixed,
    y dir=both, 
    y explicit
] table [x=x, y=y,y error=error, col sep=comma] {
    x,    y,       error
    5, 0.475, 0.041
    10, 0.616, 0.019
    50, 0.767, 0.041
    100, 0.868, 0.014

};
\addlegendentry{fasttext}

\addplot[
  black, mark=triangle*, blue, mark options={scale=3},
  error bars/.cd,
      error bar style={line width=2pt,solid},
  error mark options={line width=2pt,mark size=4pt,rotate=90},
  y fixed,
  y dir=both,
  y explicit
] table [x=x, y=y, y error=error, col sep=comma]{
  x, y, error
  5, 0.506, 0.041
  10, 0.648, 0.025
  50, 0.818, 0.008
  100, 0.862, 0.005
};
\addlegendentry{Bi-LSTM+Attn}

\addplot[
  black, mark=triangle*, cyan, mark options={scale=3},
  error bars/.cd,
    error bar style={line width=2pt,solid},
  error mark options={line width=2pt,mark size=4pt,rotate=90},
  y fixed,
  y dir=both,
  y explicit
] table [x=x, y=y, y error=error, col sep=comma]{
  x, y, error
  5, 0.350, 0.012
  10, 0.484, 0.010
  50, 0.501, 0.003
  100, 0.835, 0.005
};
\addlegendentry{HAN}

\addplot[
  black, mark=triangle*, orange, mark options={scale=3},
  error bars/.cd,
      error bar style={line width=2pt,solid},
  error mark options={line width=2pt,mark size=4pt,rotate=90},
  y fixed,
  y dir=both,
  y explicit
] table [x=x, y=y, y error=error, col sep=comma]{
  x, y, error
  5, 0.325, 0.113
  10, 0.402, 0.123
  50, 0.675, 0.050
  100, 0.787, 0.015
};
\addlegendentry{W2V}

\addplot[
  black, mark=triangle*, green, mark options={scale=3},
  error bars/.cd,
      error bar style={line width=2pt,solid},
  error mark options={line width=2pt,mark size=4pt,rotate=90},
  y fixed,
  y dir=both,
  y explicit
] table [x=x, y=y, y error=error, col sep=comma]{
  x, y, error
  5, 0.730, 0.008
  10, 0.746, 0.018
  50, 0.819, 0.008
  100, 0.829, 0.004
};
\addlegendentry{SentBERT}

\addplot[
  black, mark=triangle*, olive, mark options={scale=3},
    error bars/.cd,
    error bar style={line width=2pt,solid},
  error mark options={line width=2pt,mark size=4pt,rotate=90},
  y fixed,
  y dir=both,
  y explicit
] table [x=x, y=y, y error=error, col sep=comma]{
  x, y, error
  5, 0.964, 0.041 
  10, 0.979, 0.007
  50, 0.986, 0.002
  100, 0.987, 0.001
};
\addlegendentry{BERT}

\addplot[
  black, mark=*, black, mark options={scale=3},
  error bars/.cd,
  y fixed,
  y dir=both,
  y explicit
] table [x=x, y=y, y error=error, col sep=comma]{
  x, y, error
  5, 0.622, 0.022
  10, 0.701, 0.021
  50, 0.825, 0.003
  100, 0.857, 0.004
};
\addlegendentry{gzip}

\end{axis}
\end{tikzpicture}
\qquad
\begin{tikzpicture}[scale=0.28]
\begin{axis}[
width=1\textwidth,
height=0.80\textwidth,
legend cell align=left,
font=\LARGE,
axis y line*=left,
xmin=5, xmax=100,domain=1:10,
ymin=0.0, ymax=1.0,
every axis plot/.append style={thick},
xtick={5,10,50,100},
ytick={0.0, 0.2, 0.4, 0.6, 0.8, 1.0},
legend pos=south east,
xmajorgrids=true,
ymajorgrids=true,
legend style={nodes={scale=1.2, transform shape}},
tick label style={font=\LARGE},
xlabel style={font = \huge, yshift=0ex},
xlabel=\# of shots,
ylabel= Test Accuracy,
ylabel style={font = \huge, yshift=0ex}]

\addplot[
  black, mark=triangle*, red, mark options={scale=3},
  error bars/.cd, 
      error bar style={line width=2pt,solid},
  error mark options={line width=2pt,mark size=4pt,rotate=90},
    y fixed,
    y dir=both, 
    y explicit
] table [x=x, y=y,y error=error, col sep=comma] {
    x,    y,       error
    5, 0.545, 0.053
    10, 0.652, 0.051
    50, 0.782, 0.034
    100, 0.809, 0.012

};
\addlegendentry{fasttext}

\addplot[
  black, mark=triangle*, blue, mark options={scale=3},
  error bars/.cd,
      error bar style={line width=2pt,solid},
  error mark options={line width=2pt,mark size=4pt,rotate=90},
  y fixed,
  y dir=both,
  y explicit
] table [x=x, y=y, y error=error, col sep=comma]{
  x, y, error
  5, 0.534, 0.614
  10, 0.614, 0.047
  50, 0.771, 0.021
  100, 0.812, 0.008
};

\addlegendentry{Bi-LSTM+Attn}

\addplot[
  black, mark=triangle*, cyan, mark options={scale=3},
  error bars/.cd,
    error bar style={line width=2pt,solid},
  error mark options={line width=2pt,mark size=4pt,rotate=90},
  y fixed,
  y dir=both,
  y explicit
] table [x=x, y=y, y error=error, col sep=comma]{
  x, y, error
  5, 0.425, 0.072
  10, 0.542, 0.118
  50, 0.671, 0.102 
  100, 0.808, 0.020 
};
\addlegendentry{HAN}

\addplot[
  black, mark=triangle*, orange, mark options={scale=3},
  error bars/.cd,
      error bar style={line width=2pt,solid},
  error mark options={line width=2pt,mark size=4pt,rotate=90},
  y fixed,
  y dir=both,
  y explicit
] table [x=x, y=y, y error=error, col sep=comma]{
  x, y, error
  5, 0.141, 0.005
  10, 0.124, 0.048
  50, 0.133, 0.016
  100, 0.395, 0.089
};

\addlegendentry{W2V}

\addplot[
  black, mark=triangle*, green, mark options={scale=3},
  error bars/.cd,
      error bar style={line width=2pt,solid},
  error mark options={line width=2pt,mark size=4pt,rotate=90},
  y fixed,
  y dir=both,
  y explicit
] table [x=x, y=y, y error=error, col sep=comma]{
  x, y, error
  5, 0.485, 0.043
  10, 0.501, 0.041
  50, 0.565, 0.013
  100, 0.572, 0.003
};
\addlegendentry{SentBERT}

\addplot[
  black, mark=triangle*, olive, mark options={scale=3},
    error bars/.cd,
    error bar style={line width=2pt,solid},
  error mark options={line width=2pt,mark size=4pt,rotate=90},
  y fixed,
  y dir=both,
  y explicit
] table [x=x, y=y, y error=error, col sep=comma]{
  x, y, error
  5, 0.221, 0.041 
  10, 0.226, 0.060
  50, 0.392, 0.276
  100, 0.679, 0.073 
};
\addlegendentry{BERT}

\addplot[
  black, mark=*, black, mark options={scale=3},
  error bars/.cd,
      error bar style={line width=2pt,solid},
  error mark options={line width=2pt,mark size=4pt,rotate=90},
  y fixed,
  y dir=both,
  y explicit
] table [x=x, y=y, y error=error, col sep=comma]{
  x, y, error
  5, 0.649, 0.061
  10, 0.741, 0.017
  50, 0.833, 0.007
  100, 0.867, 0.016
};
\addlegendentry{gzip}

\end{axis}

\end{tikzpicture}%

\caption{Comparison among different methods using different shots on AG News, DBpedia and SogouNews with 95\% confidence interval over five trials.}
\label{figure:fsl}
\end{figure*}
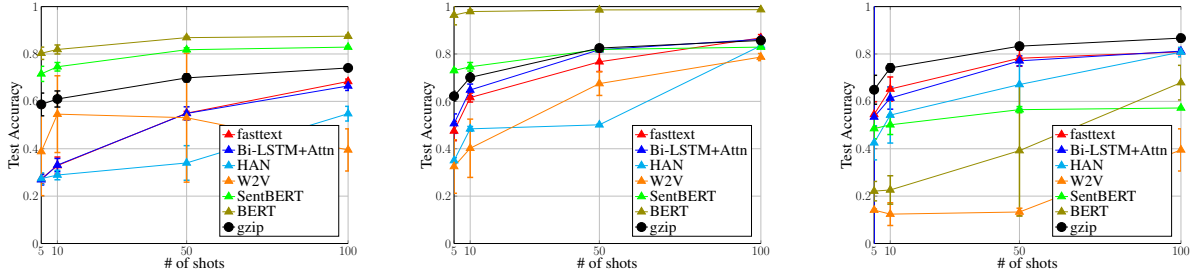

\subsection{Few-Shot Learning}
We further compare the result of \textit{gzip} under the few-shot-learning setting with deep learning methods. We first carry out experiments on AGNews, DBpedia and SogouNews across both non-pretrained deep neural networks and pre-trained ones using $n$-shot labeled examples per class from training dataset, where $n=\{5,10,50,100\}$. We chose these three datasets as their scale is large enough to cover 100-shot setting and they vary in text lengths as well as languages. We choose methods whose trainable parameters range from zero parameters like word2vec and sentence BERT to hundreds of millions of parameters like BERT, covering both word-based models (HAN) and the n-gram one (fasttext).

The result is plotted in~\Cref{figure:fsl} (detailed numbers are shown in~\Cref{appx:fsl}). As we can see, \textit{gzip} outperforms non-pretrained models on $5,10,50$ settings for all three datasets and especially in the $n=5$ setting, \textit{gzip} outperforms deep learning models by large margin. For example, the accuracy of \textit{gzip} is 115\% better than fasttext on AGNews 5-shot setting. In the 100-shot setting, \textit{gzip} also outperforms non-pretrained models on AGNews and SogouNews but is a little bit lower than them on DBpedia. 

It's been investigated in the previous work~\cite{nogueira2020document, zhang2021effectiveness} that pre-trained models are excellent few-shot learners. The advantages of BERT and SentBERT on the AGNews are obvious where they achieve the highest and the second highest accuracy on every shot number. However, on SogouNews, both BERT and SentBERT are surpassed by \textit{gzip} on every shot number, consistent with the result on full dataset. This is reasonable as the inductive bias learned from the pre-training data is so strong---notice how low the accuracy is when only given 5-shot training samples to BERT, that hinders BERT to be applied to the dataset that's significantly different from the pre-trained datasets.
The surprising part is that even on DBpedia \textit{gzip} still outperforms SentBERT on 50-shot and 100-shot settings. Note that BERT has been pre-trained on Wikipedia and DBpedia is extracted from Wikipedia, which may explain the nearly perfect score of BERT on DBpedia. In general, the larger the number of labeled training samples are, the closer that the accuracy gap between \textit{gzip} and deep learning models are, except for W2V, which is extremely unstable. This is due to the vectors being trained for a limited set of words, meaning that numerous tokens in the test set are out-of-vocabulary. 

Given pre-trained models' outstanding performance in few-shot settings on in-distributed datasets, we further investigate their few-shot performance on out-of-distributed datasets. In~\Cref{tab:ood}, we carry out experiments under 5-shot setting with BERT and mBERT. The advantage of using our method in 5-shot is more obvious than on the full datasets --- our method improves the accuracy of BERT by $90.8\%$, $40.2\%$, $59.4\%$, $58.3\%$ and $193.7\%$ and surpasses mBERT's accuracy by $100.0\%$, $67.0\%$, $40.2\%$, $12.4\%$ and $130.1\%$ on the corresponding five datasets.

\section{Analyses}
\label{sec:ana}

To understand the merits and shortcomings of using \textit{gzip} for classification, we evaluate \textit{gzip}'s performance in terms of both the absolute accuracy and the relative performance compared to the neural methods. An absolute low accuracy with a high relative performance suggests that the dataset itself is difficult, while a high accuracy with a low relative performance means the dataset is better solved by a neural network. As our method performs well on out-of-distributed datasets, we are more interested in analyzing in-distributed cases. We carry out on seven in-distributed datasets and one out-of-distributed datasets across fourteen models to account for different ranks. We analyze both the relative performance and the absolute accuracy regarding the vocabulary size and the compression rate of both datasets (i.e., how easily a dataset can be compressed) and compressors (i.e., how well a compressor can compress).

To represent the relative performance with regard to other methods, we use the normalized rank percentage, computed as $\frac{\text{rank of gzip}}{\text{total\#methods}}$; the lower the score, the better \textit{gzip} is.  We use ``bits per character''(bpc) to evaluate the compression rate. The procedure is to randomly sample a thousand instances from the training and test set respectively, calculate the compressed length and divide by the number of characters. Sampling is to 
keep the size of the dataset a constant.


\begin{figure}[th]
    \centering
\begin{tikzpicture}[scale=0.6]
\begin{axis}[
xmin=10000, xmax=1600000,
ymin=0, ymax=1.2,
every axis plot/.append style={thick},
legend pos=south east,
xmajorgrids=true,
ymajorgrids=true,
legend style={nodes={scale=0.8, transform shape}},
tick label style={font=\large},
xlabel style={font = \large, yshift=0ex},
xlabel=Vocabulary Size,
ylabel= Normalized Rank Percentage,
ylabel style={font = \large, yshift=0ex}
]

\addplot+[smooth,draw=black,mark=oplus, mark options={scale=2}] coordinates
{(128349,0.231)};
\addlegendentry{AGNews}
\addplot+[smooth,draw=black,mark=x, mark options={scale=2}] coordinates
{(1031601,0.846)};
\addlegendentry{DBpedia}
\addplot+[smooth,draw=black,  style={solid}, mark=otimes, mark options={scale=2}] coordinates
{(1554607, 1)};
\addlegendentry{YahooAnswers}
\addplot+[smooth,draw=black,mark=star, mark options={scale=2}] coordinates
{(227330, 0.538)};
\addlegendentry{20News}
\addplot+[smooth,draw=black, style={solid}, mark=diamond, mark options={scale=2}] coordinates
{(55142, 0.384)};
\addlegendentry{Ohsumed}
\addplot+[smooth,draw=black, style={solid, fill=gray}, mark=triangle, mark options={scale=2}] coordinates
{(23584, 0.231)};
\addlegendentry{R8}
\addplot+[smooth,draw=black, mark=square,style={solid}, mark options={scale=2}] coordinates
{(26283, 0.231)};
\addlegendentry{R52}
\addplot+[smooth,draw=black,mark=o, mark options={scale=2}] coordinates
{(610908, 0.077)};
\addlegendentry{SogouNews}

\end{axis}

\end{tikzpicture}
\begin{tikzpicture}[scale=0.6]
\begin{axis}[
xmin=2, xmax=3.5,
ymin=0, ymax=1.2,
every axis plot/.append style={thick},
legend pos=north west,
xmajorgrids=true,
ymajorgrids=true,
legend style={nodes={scale=0.8, transform shape}},
tick label style={font=\large},
xlabel style={font = \large, yshift=0ex},
xlabel=Bits per Character,
ylabel= Normalized Rank Percentage,
ylabel style={font = \large, yshift=0ex}
]

\addplot+[smooth,draw=black,mark=oplus, mark options={scale=2}] coordinates
{(3.03,0.231)};
\addlegendentry{AGNews}
\addplot+[smooth,draw=black,mark=x, mark options={scale=2}] coordinates
{(2.84,0.846)};
\addlegendentry{DBpedia}
\addplot+[smooth,draw=black,  style={solid}, mark=otimes, mark options={scale=2}] coordinates
{(3.31, 1)};
\addlegendentry{YahooAnswers}
\addplot+[smooth,draw=black,mark=star, mark options={scale=2}] coordinates
{(3.31, 0.538)};
\addlegendentry{20News}
\addplot+[smooth,draw=black, style={solid}, mark=diamond, mark options={scale=2}] coordinates
{(2.60, 0.384)};
\addlegendentry{Ohsumed}
\addplot+[smooth,draw=black, style={solid, fill=gray}, mark=triangle, mark options={scale=2}] coordinates
{(2.38, 0.231)};
\addlegendentry{R8}
\addplot+[smooth,draw=black, mark=square,style={solid}, mark options={scale=2}] coordinates
{(2.43, 0.231)};
\addlegendentry{R52}
\addplot+[smooth,draw=black,mark=o, mark options={scale=2}] coordinates
{(2.13, 0.077)};
\addlegendentry{SogouNews}

\end{axis}

\end{tikzpicture}

\caption{Relative performance v.s. vocabulary size and compression rate.}
\label{fig:cor}
\vspace{-1em}
\end{figure}
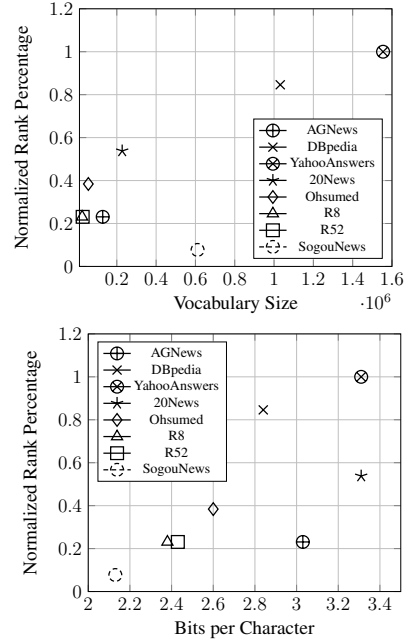

\subsection{Relative Performance}
\label{sec:rel_perf}

Combining~\Cref{tab:dataset} and~\Cref{tab:full}, we see that accuracy is largely unaffected by the average length of a single sample: with the Spearman coefficient $r_s=-0.220$. But the relative performance is more correlated with vocabulary size ($r_s=0.561$) as we can see in~\Cref{fig:cor}. SogouNews is an outlier in the first plot: on a fairly large vocabulary-sized dataset, \textit{gzip} ranks the first. The second plot may provide an explanation for that --- the compression ratio for SogouNews is high which means even with a relatively large vocabulary size, there are also repetitive information that can be squeezed out. With $r_s=0.785$ on the correlation between the normalized rank percentage and the compression rate, we can see when a dataset is easier to compress, our method may be a strong candidate as a classifier.


\begin{figure*}[th]
\centering
\begin{tikzpicture}[scale = 0.28]
\begin{axis}[
width=1\textwidth,
height=0.80\textwidth,
legend cell align=left,
font=\LARGE,
axis y line*=left,
xmin=5, xmax=100,domain=1:10,
ymin=0.5, ymax=1.0,
every axis plot/.append style={thick},
xtick={5,10,50,100},
ytick={0.0, 0.2, 0.4, 0.6, 0.8, 1.0},
legend pos=south east,
xmajorgrids=true,
ymajorgrids=true,
tick label style={font=\LARGE},
legend style={nodes={scale=1.2, transform shape}},
xlabel style={font = \huge, yshift=0ex},
xlabel=\# of shots,
ylabel= Test Accuracy,
ylabel style={font = \huge, yshift=0ex}]

\addplot[
  black, mark=*, brown, mark options={scale=3},
  error bars/.cd,
    error bar style={line width=2pt,solid},
  error mark options={line width=2pt,mark size=4pt,rotate=90},
  y fixed,
  y dir=both,
  y explicit
] table [x=x, y=y, y error=error, col sep=comma]{
  x, y, error
  5, 0.54, 0.027
  10, 0.555, 0.030
  50, 0.634, 0.040
  100, 0.667, 0.040
};
\addlegendentry{bz2}

\addplot[
  black, mark=*, violet, mark options={scale=3},
  error bars/.cd,
    error bar style={line width=2pt,solid},
  error mark options={line width=2pt,mark size=4pt,rotate=90},
  y fixed,
  y dir=both,
  y explicit
] table [x=x, y=y, y error=error, col sep=comma]{
  x, y, error
  5, 0.584, 0.036
  10, 0.601, 0.043
  50, 0.705, 0.020
  100, 0.730, 0.029
};
\addlegendentry{lzma}

\addplot[
  black, mark=*, gray, mark options={scale=3},
  error bars/.cd,
    error bar style={line width=2pt,solid},
  error mark options={line width=2pt,mark size=4pt,rotate=90},
  y fixed,
  y dir=both,
  y explicit
] table [x=x, y=y, y error=error, col sep=comma]{
  x, y, error
  5, 0.534, 0.106
  10, 0.575, 0.050
  50, 0.680, 0.016
  100, 0.739, 0.025
};
\addlegendentry{zstd}

\addplot[
  black, mark=*, black, mark options={scale=3},
  error bars/.cd,
    error bar style={line width=2pt,solid},
  error mark options={line width=2pt,mark size=4pt,rotate=90},
  y fixed,
  y dir=both,
  y explicit
] table [x=x, y=y, y error=error, col sep=comma]{
  x, y, error
  5, 0.546, 0.035
  10, 0.586, 0.036
  50, 0.718, 0.009
  100, 0.769, 0.012
};
\addlegendentry{gzip}

\end{axis}
\end{tikzpicture}%
\qquad
\begin{tikzpicture}[scale = 0.28]
\begin{axis}[
width=1\textwidth,
height=0.80\textwidth,
legend cell align=left,
font=\LARGE,
axis y line*=left,
xmin=5, xmax=100,domain=1:10,
ymin=0.5, ymax=1.0,
every axis plot/.append style={thick},
xtick={5,10,50,100},
ytick={0.0, 0.2, 0.4, 0.6, 0.8, 1.0},
legend pos=south east,
xmajorgrids=true,
ymajorgrids=true,
tick label style={font=\LARGE},
legend style={nodes={scale=1.2, transform shape}},
xlabel style={font = \huge, yshift=0ex},
xlabel=\# of shots,
ylabel= Test Accuracy,
ylabel style={font = \huge, yshift=0ex}]

\addplot[
  black, mark=*, brown, mark options={scale=3},
  error bars/.cd,
    error bar style={line width=2pt,solid},
  error mark options={line width=2pt,mark size=4pt,rotate=90},
  y fixed,
  y dir=both,
  y explicit
] table [x=x, y=y, y error=error, col sep=comma]{
  x, y, error
  5, 0.605, 0.057
  10, 0.698, 0.042
  50, 0.798, 0.012
  100, 0.855, 0.014
};
\addlegendentry{bz2}

\addplot[
  black, mark=*, violet, mark options={scale=3},
  error bars/.cd,
    error bar style={line width=2pt,solid},
  error mark options={line width=2pt,mark size=4pt,rotate=90},
  y fixed,
  y dir=both,
  y explicit
] table [x=x, y=y, y error=error, col sep=comma]{
  x, y, error
  5, 0.590, 0.104
  10, 0.621, 0.029
  50, 0.786, 0.038
  100, 0.823, 0.024
};
\addlegendentry{lzma}

\addplot[
  black, mark=*, gray, mark options={scale=3},
  error bars/.cd,
    error bar style={line width=2pt,solid},
  error mark options={line width=2pt,mark size=4pt,rotate=90},
  y fixed,
  y dir=both,
  y explicit
] table [x=x, y=y, y error=error, col sep=comma]{
  x, y, error
  5, 0.629, 0.036
  10, 0.683, 0.037
  50, 0.820, 0.017
  100, 0.855, 0.012
};
\addlegendentry{zstd}

\addplot[
  black, mark=*, black, mark options={scale=3},
  error bars/.cd,
      error bar style={line width=2pt,solid},
  error mark options={line width=2pt,mark size=4pt,rotate=90},
  y fixed,
  y dir=both,
  y explicit
] table [x=x, y=y, y error=error, col sep=comma]{
  x, y, error
  5, 0.648, 0.048
  10, 0.676, 0.020
  50, 0.822, 0.013
  100, 0.849, 0.011
};
\addlegendentry{gzip}

\end{axis}
\end{tikzpicture}%
\qquad
\begin{tikzpicture}[scale = 0.28]
\begin{axis}[
width=1\textwidth,
height=0.80\textwidth,
legend cell align=left,
font=\LARGE,
axis y line*=left,
xmin=5, xmax=100,domain=1:10,
ymin=0.5, ymax=1.0,
every axis plot/.append style={thick},
xtick={5,10,50,100},
ytick={0.0, 0.2, 0.4, 0.6, 0.8, 1.0},
legend pos=south east,
xmajorgrids=true,
ymajorgrids=true,
tick label style={font=\LARGE},
legend style={nodes={scale=1.2, transform shape}},
xlabel style={font = \huge, yshift=0ex},
xlabel=\# of shots,
ylabel= Test Accuracy,
ylabel style={font = \huge, yshift=0ex}]

\addplot[
  black, mark=*, brown, mark options={scale=3},
  error bars/.cd,
    error bar style={line width=2pt,solid},
  error mark options={line width=2pt,mark size=4pt,rotate=90},
  y fixed,
  y dir=both,
  y explicit
] table [x=x, y=y, y error=error, col sep=comma]{
  x, y, error
  5, 0.546, 0.035
  10, 0.586, 0.036
  50, 0.718, 0.009
  100, 0.769, 0.012
};
\addlegendentry{bz2}

\addplot[
  black, mark=*, violet, mark options={scale=3},
  error bars/.cd,
    error bar style={line width=2pt,solid},
  error mark options={line width=2pt,mark size=4pt,rotate=90},
  y fixed,
  y dir=both,
  y explicit
] table [x=x, y=y, y error=error, col sep=comma]{
  x, y, error
  5, 0.605, 0.045
  10, 0.683, 0.024
  50, 0.827, 0.007
  100, 0.860, 0.009
};
\addlegendentry{lzma}

\addplot[
  black, mark=*, gray, mark options={scale=3},
  error bars/.cd,
    error bar style={line width=2pt,solid},
  error mark options={line width=2pt,mark size=4pt,rotate=90},
  y fixed,
  y dir=both,
  y explicit
] table [x=x, y=y, y error=error, col sep=comma]{
  x, y, error
  5, 0.569, 0.039
  10, 0.621, 0.043
  50, 0.764, 0.023
  100, 0.812, 0.019
};
\addlegendentry{zstd}

\addplot[
  black, mark=*, black, mark options={scale=3},
  error bars/.cd,
  y fixed,
  y dir=both,
  y explicit
] table [x=x, y=y, y error=error, col sep=comma]{
  x, y, error
  5, 0.629, 0.025
  10, 0.692, 0.021
  50, 0.828, 0.013
  100, 0.857, 0.011
};
\addlegendentry{gzip}

\end{axis}
\end{tikzpicture}
\caption{Comparison among different compressors on AG News, SogouNews and DBpedia, with 95\% confidence interval over five trials.}
\label{figure:fsl-cp}
\end{figure*}
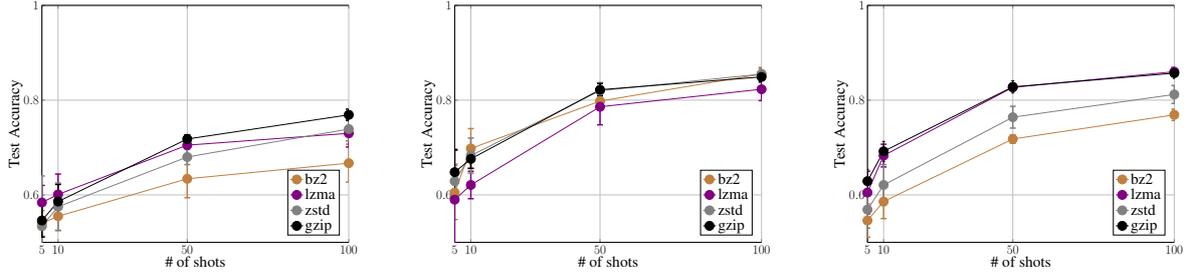

\begin{figure}[h!]
    \centering
    \includegraphics[width=.9\linewidth]{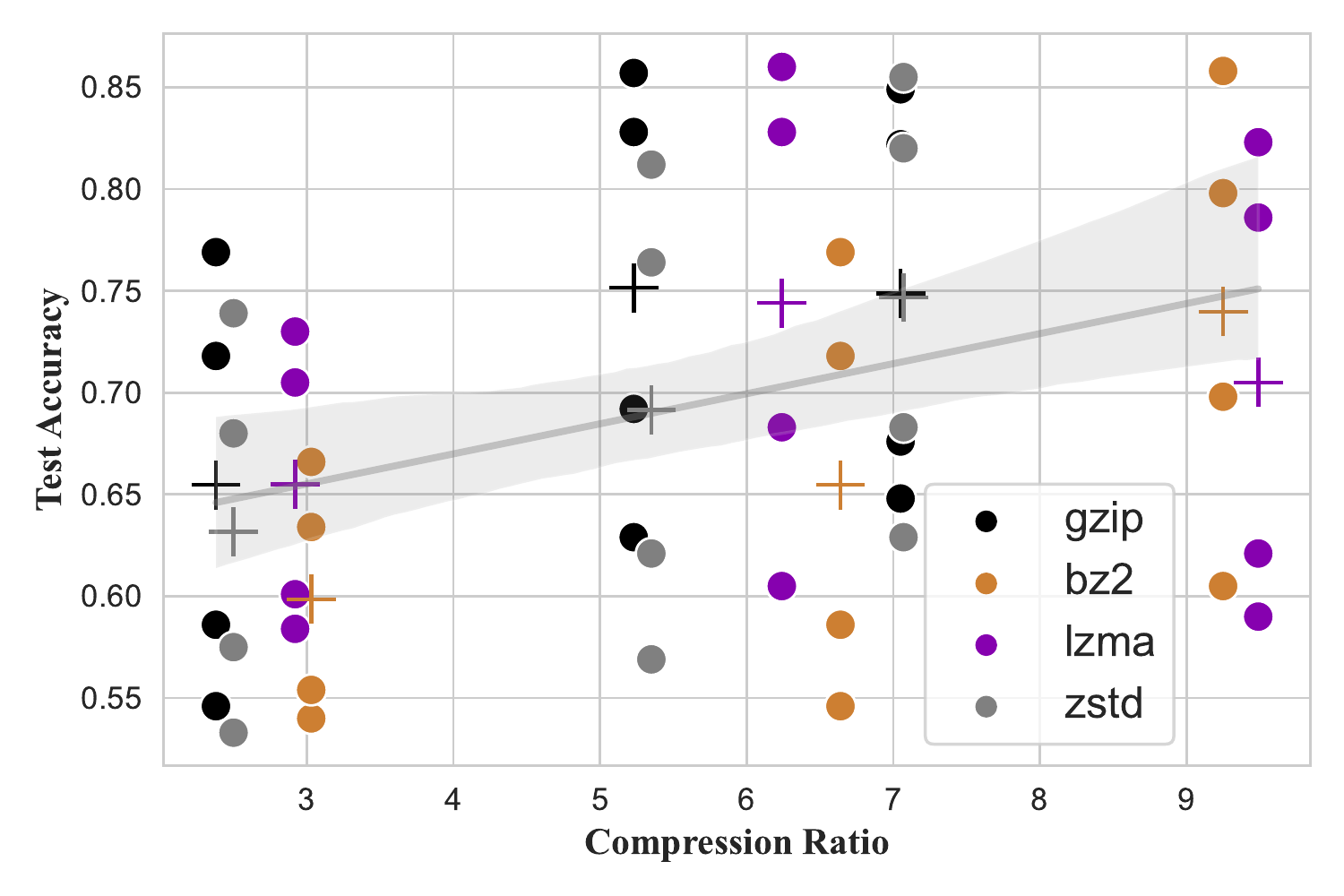}
    \caption{Compression ratio V.S. Test Accuracy across different compressors on three datasets under different shot settings}
    \label{fig:ratio_acc}
\end{figure}

\subsection{Absolute Accuracy}
Similarly we evaluate the accuracy of classification with respect to the vocabulary size and we've found there is almost no monotonic relation ($r_s=0.071)$. With regard to bpc, the monotonic relation is not as strong as the one with the rank percentage ($r_s=-0.56$). Considering the effect that vocabulary size has on the relative performance, our method with \textit{gzip} may be more susceptible to the vocabulary size than neural network methods. To distinguish between a ``hard'' dataset and an ``easy'' one, we average all models' accuracies. The dataset that has the lowest accuracies are 20News and Ohsumed, which are two datasets that have the longest average length of texts. 

\subsection{Using Other Compressors}
With compressor-based distance metrics we can use any compressor. Because of the large size of the test set of the datasets, we randomly chose 1,000 test samples to evaluate and repeat the experiments for each setting five times to calculate the mean and 95\% confidence interval. 

We carry out experiments on other three compressors: \textit{bz2}, \textit{lzma} and \textit{zstandard} under the few-shot setting. Each of them has different underlying algorithms from \textit{gzip}. \textit{bz2} uses Burrows-Wheeler algorithm to permute the order of characters in the strings to create more repeated ``substrings'' that can be compressed. That's one of the reasons why \textit{bz2} has a higher compression ratio (e.g., it can achieve 2.57 bpc on AGNews while \textit{gzip} can achieve only 3.38 bpc). \textit{lzma} is based on LZ77, a dictionary-based compression algorithm, where the idea is to use \textit{(offset, length)} to represent the n-gram that has previously appeared in the search buffer. \textit{lzma} then uses range coding to further encode \textit{(offset, length)}. Similarly, \textit{gzip} uses DEFLATE algorithm, which also uses LZ77 and instead of range coding, it takes advantage of Huffman coding to further encode \textit{(offset, length)}. \textit{zstandard (zstd)} is a new compression algorithm that's built on LZ77, Huffman coding as well as Asymmetric Numeral Systems (ANS)~\cite{duda2009asymmetric}. We pick \textit{zstd} to evaluate for its fast speed, with close compression rate to \textit{gzip}. A competitive result may indicate it can be used to speed up the classification. 

We plot all the test accuracy in~\Cref{fig:ratio_acc} with the compression ratio for each compressor. Compression ratio is calculated by $\frac{\text{original size}}{\text{compressed size}}$, so the larger the compression ratio is, the more a compressor can compress. We use compression ratio instead of bpc here as the latter one is too close to each other and cannot be differentiated from one another. Markers of `+' represents the mean of each compressor's test accuracy across different shot settings. The dataset is not explicitly labeled but we can tell that there are roughly three clusters in the plot. AGNews is the cluster with the lowest compression ratio and SogouNews is the one with the highest compression ratio. Note that \textit{gzip} and \textit{zstd} with compression ratio of about 7 belongs to the SogouNews result. 

On SogouNews, both \textit{gzip} and \textit{zstd} have the compression ratio equal to about 7; \textit{bz2} and \textit{lzma} have the compression ratio over 9. The difference of accuracy is more obvious on the AGNews and DBpedia with \textit{bz2} being the worst-performing compressor. This is counterintuitive, as a compressor with a higher compression ratio suggests that it can approximate Kolmogorov complexity better, and \textit{bz2} has a higher compression ratio. We conjecture it may be because in practice, Burrows-Wheeler algorithm used by \textit{bz2} dismisses the information of character order. This is shown more clearly in~\Cref{fig:ratio_acc} --- \textit{bz2} is always lower than the regression line. In general, \textit{gzip} achieves a relatively high and stable accuracy across three datasets. \textit{lzma} is competitive with \textit{gzip} but the speed is much slower.

We've found in~\Cref{sec:rel_perf} that for a single compressor, the easier a dataset can compress, the more probable it can achieve a higher accuracy than deep learning models. Here we investiage the correlation across compressors. We've found the compression ratio and test accuracy has a moderate monotonic and linear correlation and as the shot number increases, the linear correlation is more obvious with $r_s=0.605$ for all shot settings and Pearson correlation $r_p=0.575, 0.638, 0.691, 0.719$ respectively on 5, 10, 50 and 100 setting across four compressors. Combining the special case of \textit{bz2} with the linear correlation between compression ratio and test accuracy, we know that in general a compressor with a high compression ratio can perform better on a more compressible dataset. But the actual compression algorithm still has its effect on the test accuracy despite the high compression ratio.

\begin{table}[t]
\resizebox{\linewidth}{!}{%
    \centering
    {\renewcommand{\arraystretch}{1.2}
    \begin{tabular}{c|cccc}
        \scalebox{1.3}{Method} & \scalebox{1.3}{AGNews} & \scalebox{1.3}{SogouNews} & \scalebox{1.3}{DBpedia} & \scalebox{1.3}{YahooAnswers} \\
        \toprule
        \scalebox{1.2}{gzip(ce)} & \scalebox{1.3}{0.739\sbpm{0.046}} & \scalebox{1.3}{0.741\sbpm{0.076}} & \scalebox{1.3}{0.880\sbpm{0.010}} & \scalebox{1.3}{0.408\sbpm{0.012}} \\
        \hline
        \scalebox{1.2}{gzip($k$NN)} & \scalebox{1.3}{0.752\sbpm{0.041}} & \scalebox{1.3}{0.862\sbpm{0.033}} & \scalebox{1.3}{0.852\sbpm{0.008}} & \scalebox{1.3}{0.352\sbpm{0.014}} \\
        \bottomrule
    \end{tabular}
    }
    }
    \caption{Comparison with other compressor-based methods under the 100-shot setting.}
    \label{tab:comp2}
    \vspace{-1em}
\end{table}

\subsection{Using Other Compressor-Based Methods}
The distance metric used by previous work~\cite{marton2005compression, russell2010artificial} is mainly $C(d_cd_u)-C(d_c)$ as we mention in~\Cref{sec:ctc}.
Although using this distance metric is faster than pair-wise distance matrix computation on small datasets, it has several drawbacks: (1) Most compressors have a limited ``size'', for \textit{gzip} it's the sliding window size that can be used to search back of the repeated string while for \textit{lzma} it's the dictionary size it can keep record of. This means even if there are large number of training samples, the compressor cannot take full advantage of those samples; (2) When $d_c$ is large, compressing $d_cd_u$ can be really slow and this slowness cannot be solved by parallelization. These two main drawbacks stop this method to be applied to a really large dataset. Thus, we randomly pick 1000 test samples and 100-shot from each class in training samples to compare these two methods. In~\Cref{tab:comp2}, ``\textit{gzip} (ce)'' means using the cross entropy $C(d_cd_u)-C(d_c)$ while ``\textit{gzip} ($k$NN)'' refers to our method. We carry our each experiment for five times and calculate the mean and 95\% confidence interval. On AGNews and SogouNews using $k$NN and NCD is better than using cross entropy. The reason for the large accuracy gap between them on SogouNews is probably because each instance in SogouNews is very long, causing about 11.2K per sample, while \textit{gzip} typically has 32K window size only. Only concatenation a few samples makes the compression ineffective. The cross-entropy method does perform very well on YahooAnswers, which may benefit from using multiple references in the single category as YahooAnswers is a divergent dataset created by numerous online users.

We also test the performance of compressor-based cross entropy method on \textit{full} AGNews dataset as it is a relatively smaller one with shorter single instance. The accuracy is 0.745, not much higher than 100-shot setting, which further confirms that using $C(d_cd_u)-C(d_c)$ as a distance metric cannot take full advantage of the large datasets.

\section{Conclusions and Future Work}
In this paper, we use \textit{gzip} together with a compressor-based distance metric to achieve classification accuracy comparable to neural network classifiers on in-distributed datasets and outperform pre-trained models on out-of-distributed datasets. We also show the effectiveness of using this method in few-shot scenarios.
In future works, we will extend this work by generalizing \textit{gzip} to neural compressors on text, as recent studies~\cite{jiang2022few} show that combining neural compressors that derived from deep latent variables models with compressor-based distance metrics for image classification can even outperform semi-supervised methods.

\bibliography{anthology, papers}
\bibliographystyle{acl_natbib}

\newpage

\appendix

\section{Derivation of NCD}
\label{appx:ncd}
Recall that \textit{information distance} $E(x,y)$ is:

\begin{align}
            E(x,y) & = \max\{K(x|y), K(y|x)\} \\
    & = K(xy) - \min \{K(x), K(y)\}
\end{align}
$E(x,y)$ equates the similarity between two objects with the existence of a program that can convert one to another. The simpler the converting program is, the more similar the objects are. For example, the negative of an image is very similar to the original one as the transformation can be simply described as ``inverting the color of the image''.

In order to compare the similarity, relative distance is preferred. \citet{vitanyi2009normalized} propose a normalized version of $E(x,y)$ called \textit{Normalized Information Distance} (NID). 
\begin{definition}[\textbf{NID}]
NID is a function: $ \Omega\times\Omega \rightarrow [0,1]$, where $\Omega$ is a non-empty set, defined as:
\begin{equation}
    \text{NID}(x,y) = \frac{\max \{K(x|y), K(y|x)\}}{\max \{K(x), K(y)\}}.
    \label{nid}
\end{equation}
\end{definition}
\Cref{nid} can be interpreted as follows: Given two sequences $x$, $y$, $K(y)\geq K(x)$:
\begin{equation}
    \text{NID}(x,y) = \frac{K(y)-I(x:y)}{K(y)} = 1 - \frac{I(x:y)}{K(y)},
    \label{nid2}
\end{equation}
where $I(x:y)=K(y)-K(y|x)$ means the \textit{mutual algorithmic information}. $\frac{I(x:y)}{K(y)}$ means the shared information (in bits) per bit of information contained in the most informative sequence, and ~\Cref{nid2} here is a specific case of~\Cref{nid}. 

\textit{Normalized Compression Distance} (NCD) is a computable version of NID based on real-world compressors. In this context, $K(x)$ can be viewed as the length of $x$ after being maximally compressed. Suppose we have $C(x)$ as the length of compressed $x$ produced by a real-world compressor, then NCD is defined as:
\begin{equation}
    \text{NCD}(x, y) = \frac{C(xy) - \min \{C(x), C(y)\}}{\max\{C(x), C(y)\}}.
    \label{ncd}
\end{equation}

NCD is thus computable in that it not only uses compressed length to approximate $K(x)$ but also replaces conditional Kolmogorov complexity with $C(xy)$ that only needs a simple concatenation of $x,y$.

\section{Implementation Details}
\label{appx:train}

We use different hyper-parameters for full-dataset setting and few-shot setting. 

For both LSTM, Bi-LSTM+Attn, fasttext, we use embedding size $=256$, dropout rate $=0.3$. For full-dataset setting, the learning rate is set to be $0.001$ and decay rate $=0.9$ for Adam optimizer~\cite{kingma2015adam}, number of epochs $=20$, with batch size $=64$; for few-shot setting, the learning rate $=0.01$, the decay rate $=0.99$, batch size $=1$, number of epochs $=50$ for 50-shot and 100-shot, epoch $=80$ for 5-shot and 10-shot. For LSTM and Bi-LSTM+Attn, we set RNN layer $=1$, hidden size $=64$. For fasttext, we use 1 hidden layer whose dimension is set to be 10. 

For HAN, we use 1 layer for both word-level RNN and sentence-level RNN, the hidden size of both of them are set to 50, the hidden sizes of both attention layers are set to be 100. It's trained with batch size $=256$, $0.5$ decay rate for $6$ epochs. 

For BERT, the learning rate is set to be $2e-5$ and the batch size is set to be $128$ for English and SogouNews while for low-resource languages, we set learning rate to be $1e-5$ with batch size to be 16 for 5 epochs. We use transformers library for BERT's implementation and specifically we use \texttt{bert-base-uncased} checkpoint for BERT and \texttt{bert-base-multilingual-uncased} for mBERT.

For charCNN and textCNN, we use the same hyper-parameters setting in~\citet{adhikari2019rethinking} except when in the few-shot learning setting, we reduce the batch size to $1$, reducing the learning rate to $1e-4$ and increase the number of epochs to $60$.
For VDCNN, we use the shallowest $9$-layer version with embedding size set to be $16$, batch size set to be $64$ learning rate set to be $1e-4$ for full-dataset setting and batch size $=1$, epoch number $=60$ for few-shot setting. For RCNN, we use embedding size $=256$, hidden size of RNN $=256$, learning rate $=1e-3$ and same batch size and epoch setting as VDCNN for full-dataset and few-shot settings.

For pre-processing, we don't use any pre-trained word embedding for any word-based models. Neither do we use data augmentation during the training. The procedures of tokenization for both word-level and character-level, padding for batch processing are, however, inevitable.

For all zero-training methods, the only hyper-parameter is $k$.
We set $k=2$ for all the methods on all the datasets and we report the maximum possible accuracy getting from the experiments for each method. For Sentence-BERT, we use the ``paraphrase-MiniLM-L6-v2'' checkpoint.

For neural network methods, we use publicly available code for charCNN and textCNN implemented by~\citet{adhikari2019rethinking}, and we use~\citet{wolf2020transformers} for BERT. 

Our method only requires CPUs and we use 8-core CPUs to take advantage of multi-processing. The time of calculating distance matrix using \textit{gzip} takes about half an hour on AGNews, two days on DBpedia and SogouNews, six days on YahooAnswers.

All the datasets can be downloaded from \href{https://pytorch.org/text/stable/index.html}{torchtext}, \href{http://disi.unitn.it/moschitti/corpora.htm}{text categorization corpora} and hugging face datasets (\href{https://huggingface.co/datasets/kinnews_kirnews}{Kinyarwanda and Kirundi News}, \href{https://huggingface.co/datasets/swahili_news}{Swahili News}, \href{https://huggingface.co/datasets/dengue_filipino}{Dengue Filipino}).


\section{Few-Shot Results}
\label{appx:fsl}
The exact numerical value of accuracy shown in~\Cref{figure:fsl} is listed in three tables below.

\begin{table}[H]
\begin{small}
    \centering
    \resizebox{\linewidth}{!}{%
    \begin{tabular}{c|cccc}
         Dataset & \multicolumn{4}{c}{AGNews} \\
         \hline
         \#Shot & 5 & 10 & 50 &100 \\
        \hline
        fasttext & 0.273\sbpm{0.021} & 0.329\sbpm{0.036} & 0.550\sbpm{0.008} & 0.684\sbpm{0.010} \\
        \hline
        Bi-LSTM+Attn & 0.269\sbpm{0.022} & 0.331\sbpm{0.028} & 0.549\sbpm{0.028} & 0.665\sbpm{0.019} \\
         \hline
         HAN & 0.274\sbpm{0.024} & 0.289\sbpm{0.020} & 0.340\sbpm{0.073} & 0.548\sbpm{0.031} \\
         \hline
         W2V & 0.388\sbpm{0.186} & 0.546\sbpm{0.162} & 0.531\sbpm{0.272} & 0.395\sbpm{0.089} \\
         \hline
         BERT & 0.803\sbpm{0.026} & 0.819\sbpm{0.019} & 0.869\sbpm{0.005} & 0.875\sbpm{0.005} \\
         \hline
         SentBERT & 0.716\sbpm{0.032} & 0.746\sbpm{0.018} & 0.818\sbpm{0.008} & 0.829\sbpm{0.004} \\
         \hline
         gzip & 0.587\sbpm{0.048} & 0.610\sbpm{0.034} & 0.699\sbpm{0.017} & 0.741\sbpm{0.007} \\
         \hline
    
    \end{tabular}
    }
    \caption{Few-Shot result on AG News}
    \label{tab:my_label}
\end{small}
\end{table}

\begin{table}[H]
\begin{small}
    \centering
    \resizebox{\linewidth}{!}{%
    \begin{tabular}{c|cccc}
         Dataset & \multicolumn{4}{c}{DBpedia} \\
                  \hline
         \#Shot & 5 & 10 & 50 &100 \\
                  \hline
        fasttext & 0.475\sbpm{0.041} & 0.616\sbpm{0.019} & 0.767\sbpm{0.041} & 0.868\sbpm{0.014} \\
         \hline
        Bi-LSTM+Attn & 0.506\sbpm{0.041} & 0.648\sbpm{0.025} & 0.818\sbpm{0.008} & 0.862\sbpm{0.005} \\
        \hline
        HAN & 0.350\sbpm{0.012} & 0.484\sbpm{0.010} & 0.501\sbpm{0.003} & 0.835\sbpm{0.005} \\
         \hline
         W2V & 0.325\sbpm{0.113} & 0.402\sbpm{0.123} & 0.675\sbpm{0.05} & 0.787\sbpm{0.015} \\
         \hline
        BERT & 0.964\sbpm{0.041} & 0.979\sbpm{0.007} & 0.986\sbpm{0.002} & 0.987\sbpm{0.001} \\
        \hline
         SentBERT & 0.730\sbpm{0.008} & 0.746\sbpm{0.018} & 0.819\sbpm{0.008} & 0.829\sbpm{0.004} \\
         \hline
         gzip & 0.622\sbpm{0.022} & 0.701\sbpm{0.021} & 0.825\sbpm{0.003} & 0.857\sbpm{0.004} \\
         \hline
    \end{tabular}
    }
    \caption{Few-Shot result on DBpedia}
    \label{tab:my_label}
\end{small}
\end{table}

\begin{table}[H]
\begin{small}
    \centering
    \resizebox{\linewidth}{!}{%
    \begin{tabular}{c|cccc}
         Dataset & \multicolumn{4}{c}{SogouNews} \\
        \hline
         \#Shot & 5 & 10 & 50 &100 \\
        \hline
        fasttext & 0.545\sbpm{0.053} & 0.652\sbpm{0.051} & 0.782\sbpm{0.034} & 0.809\sbpm{0.012} \\ 
         \hline
        Bi-LSTM+Attn & 0.534\sbpm{0.042} & 0.614\sbpm{0.047} & 0.771\sbpm{0.021} & 0.812\sbpm{0.008} \\
        \hline
        HAN & 0.425\sbpm{0.072} & 0.542\sbpm{0.118} & 0.671\sbpm{0.102} & 0.808\sbpm{0.020} \\
         \hline
         W2V & 0.141\sbpm{0.005} & 0.124\sbpm{0.048} & 0.133\sbpm{0.016} & 0.395\sbpm{0.089} \\
        \hline
        BERT & 0.221\sbpm{0.041} & 0.226\sbpm{0.060} & 0.392\sbpm{0.276} & 0.679\sbpm{0.073} \\
        \hline
         SentBERT & 0.485\sbpm{0.043} & 0.501\sbpm{0.041} & 0.565\sbpm{0.013} & 0.572\sbpm{0.003} \\
         \hline
         gzip & 0.649\sbpm{0.061} & 0.741\sbpm{0.017} & 0.833\sbpm{0.007} & 0.867\sbpm{0.016}\\
         \hline
    \end{tabular}
    }
    \caption{Few-Shot result on SogouNews}
    \label{tab:my_label}
\end{small}
\end{table}

\end{document}